# Bisimulation-based Approximate Lifted Inference

**Prithviraj Sen**
Computer Science Department
Univ. of Maryland, College Park
sen@cs.umd.edu

**Amol Deshpande**
Computer Science Department
Univ. of Maryland, College Park
amol@cs.umd.edu

**Lise Getoor**
Computer Science Department
Univ. of Maryland, College Park
getoor@cs.umd.edu

## Abstract

There has been a great deal of recent interest in methods for performing lifted inference; however, most of this work assumes that the first-order model is given as input to the system. Here, we describe lifted inference algorithms that determine symmetries and automatically *lift* the probabilistic model to speedup inference. In particular, we describe approximate lifted inference techniques that allow the user to trade off inference accuracy for computational efficiency by using a handful of tunable parameters, while keeping the error bounded. Our algorithms are closely related to the graph-theoretic concept of bisimulation. We report experiments on both synthetic and real data to show that in the presence of symmetries, run-times for inference can be improved significantly, with approximate lifted inference providing orders of magnitude speedup over ground inference.

## 1 Introduction

While recent work in lifted inference [3, 15, 16, 17, 20] are promising steps towards developing efficient inference algorithms that can exploit the first-order structure provided by most first-order probabilistic models (see [8] for a survey), all of these techniques assume the first-order structure is provided in the input. In this work, we study the alternate problem of identifying the symmetry present in an underlying probabilistic model, and show how this can be exploited to provide new lifted inference algorithms.

Our work builds on recent results on efficient query evaluation in probabilistic databases from the database community. A query over a probabilistic database results in a very large graphical model which has many repeated factors. In prior work [18], we showed how, using the symmetry present in the probabilistic model and methods closely related to the graph-theoretic concept of bisimulation, it is possible to compile a compressed version of the inference problem. The compressed data-structure, called an *rv-elim graph*, can then be used to perform faster inference. In this paper, we show how the above techniques are generally applicable to arbitrary graphical models, and, more importantly, develop approximate lifted inference techniques that allow the user to trade off accuracy of inference for computational efficiency. We show how our approximate lifted inference techniques can compress the rv-elim graph well beyond the compression achieved by exact lifted inference, producing more impressive speedups while keeping the error bounded using certain tunable parameters.

Here is a summary of our contributions and results:

- We review the results from [18] and show how they are applicable to general probabilistic graphical models.

- Using techniques based on approximate bisimulation, we extend these methods and introduce a tunable parameter to move from approximate inference with high speedups to exact inference with perfect accuracy.

- We introduce a second approximation method to bin factors (or clique potentials) that are within a user-specified $\varepsilon$ distance of each other into common partitions. Using these partitions, it is possible to compress the rv-elim graph and perform even faster inference.

- We also show how to integrate our techniques with existing bounded complexity inference techniques (e.g., mini-buckets [5]) – that allows us to extend the use of our techniques to domains with unbounded treewidth.

- We discuss how to integrate all of the above techniques into one single inference engine that allows combinations ranging from exact lifted inference to approximate inference based on approximate bisimulation and factor binning with the use of mini-buckets.

- We experiment with synthetic and real-world data, and demonstrate how our techniques can achieve significant speedups of upto two orders of magnitude over ground inference and exact lifted inference with bounded error.

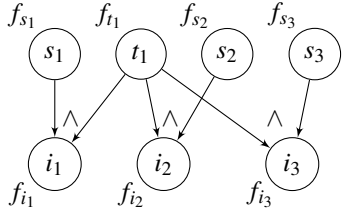

Figure 1: An example probabilistic model.

In the next section, we review our earlier work on exact lifted inference [18], in Section 3 we present our techniques for approximate lifted inference, in Section 4 we evaluate our approaches on synthetic and real-world data, in Section 5 we review related work and we conclude with a few pointers for future work in Section 6.

## 2 Background: Exact Lifted Inference with the RV-Elim Graph

This section reviews material from [18]. We begin with some notation. Let $X$ denote a random variable that can be assigned a value from a pre-defined domain denoted by $dom(X)$. Let $f(\mathbf{X})$ denote a factor or clique potential that takes as arguments a set of random variables $\mathbf{X} = \{X_1, \ldots X_n\}$. $f(\mathbf{X})$ denotes a mapping $f : dom(X_1) \times \ldots \times dom(X_n) \to \Re_{\geq 0}$. Given a set of such factors, $\mathscr{F} = \{f_1, f_2, \ldots f_m\}$, we can define a joint probability distribution over a set of random variables $\mathscr{X}$ by multiplying all the factors $Pr(\mathscr{X}) = \frac{1}{Z} \prod_{f \in \mathscr{F}} f(\mathbf{X}_f)$ such that $\mathbf{X}_f \subseteq \mathscr{X}$ denotes the arguments of $f \in \mathscr{F}$ and $\mathscr{X}$ denotes the partition function. Given such a joint probability distribution and a random variable $X \in \mathscr{X}$, let $\mu(X)$ denote its marginal probability distribution such that $\mu(X) = \sum_{\mathscr{X} \setminus X} Pr(\mathscr{X})$.

In first-order probabilistic models, many factors come from grounding out first-order rules. The factors obtained from such rules map the same input to the same outputs and constitute symmetry in the model. Consider the *friends and smokers* domain where we want to infer the probability of a person being a smoker. Then a rule such as $\texttt{Smoker}(P_1) \land \texttt{Friend}(P_1, P_2) \Rightarrow \texttt{Smoker}(P_2)$ would ground out to provide $\binom{n}{2}$ factors with identical input-output mappings, assuming all pairs $P_1, P_2$ are friends. The notion of shared factors precisely captures this symmetry.

We refer to two factors $f$ and $f'$ as being *shared*, denoted $f \cong f'$, if they contain the same input-output mappings, irrespective of whether they take the same random variables as arguments. More precisely, let $\mathbf{X} = \{X_1, X_2, \ldots X_n\}$ and $\mathbf{X'} = \{X'_1, X'_2, \ldots X'_n\}$ denote the arguments of $f$ and $f'$ respectively. Then, $f(\mathbf{X}) \cong f'(\mathbf{X'})$ iff $dom(X_i) = dom(X'_i), \forall i = 1 \ldots n$, and $f(\mathbf{x}) = f'(\mathbf{x}), \forall \mathbf{x} \in dom(X_1) \times \ldots dom(X_n)$.

Consider the small example shown in Figure 1. All random variables are boolean valued. The priors on $s_1$, $s_2$,

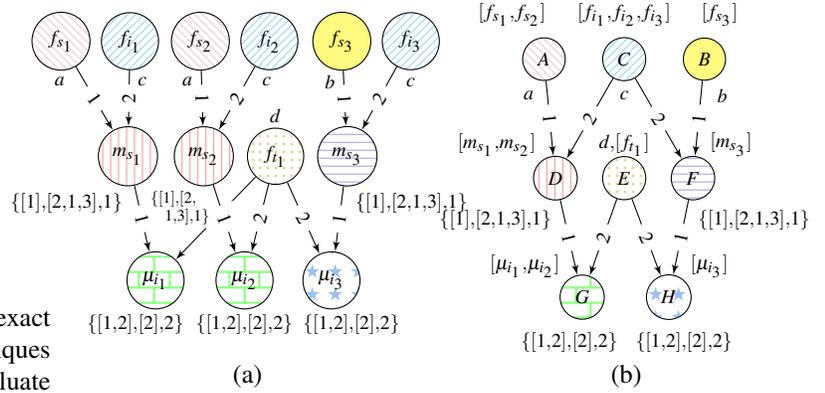

Figure 2: (a) RV-Elim graph for the running example (vertices partitioned into 8 blocks, shading indicates partitioning), (b) corresponding compressed rv-elim graph.

$s_3$ and $t_1$, denoted by $f_{s_1}$, $f_{s_2}$, $f_{s_3}$ and $f_{t_1}$, respectively, are such that $s_1$ and $s_2$ are $\texttt{true}$ with probability 0.8, $s_3$ is $\texttt{true}$ with probability 0.6 and $t_1$ is $\texttt{true}$ with probability 0.5. The three random variables $i_j, \forall j \in \{1, 2, 3\}$ are each $\texttt{true}$ when their corresponding parents $s_j$ and $t_1$ are both $\texttt{true}$ and this is enforced by the three factors $f_{i_j}(i_j, s_j, t_1)$ which return 1 iff $i_j \Leftrightarrow s_j \land t_1$ and return 0 otherwise. Of the four priors, $f_{s_1}(s_1)$ and $f_{s_2}(s_2)$ represent a pair of shared factors since they contain the same input-output mappings ($f_{s_1}(\texttt{true}) = f_{s_2}(\texttt{true}) = 0.8$); similarly, $f_{i_1}, f_{i_2}$ and $f_{i_3}$ are also shared since they are all $\texttt{and}$ factors.[*]

Given a set of factors $\mathscr{F}$, a set of random variables $\mathbf{X}$ whose marginal probabilities we are interested in computing, and an elimination order $\mathscr{O}$ which contains all the random variables to be summed over from $\mathscr{F}$ (we discuss how to construct elimination orders subsequently), it is straightforward to construct an *rv-elim* graph $G = (V, E, \mathscr{L}_V, \mathscr{L}_E)$ which is a directed acyclic graph (DAG) with vertex labels $\mathscr{L}_V$ and edge labels $\mathscr{L}_E$ such that:

- If $v$ is a root then it represents a factor $f$ from $\mathscr{F}$ and its label $\mathscr{L}_V(v)$ is such that $\forall f' \in \mathscr{F}, f' \cong f \Leftrightarrow \mathscr{L}_V(v') = \mathscr{L}_V(v)$ where $v'$ denotes the vertex representing $f'$.

- If $v$ is an internal vertex, then it represents an intermediate factor created during inference, denoting a variable elimination (summing over) operation formed by multiplying the factors represented by its parent vertices. The edge labels on edges with $v$ as head denote the order in which the parents were multiplied and the label on $v$ denotes how the arguments of its parents overlap.

In Figure 2(a), we show the rv-elim graph generated for our running example using the elimination order $\mathscr{O} = \{t_1, s_3, s_2, s_1\}$ (variables on the right are eliminated first). The vertex labels are shown next to each vertex. Notice that $f_{s_1}$ and $f_{s_2}$ have been assigned the same label "a" but $f_{s_3}$ has been assigned a different label "b"; similarly, $f_{i_1}$, $f_{i_2}$ and $f_{i_3}$ have been assigned the same label "c". Each internal vertex corresponds to an elimination operation. For instance, $m_{s_1}$ denotes the intermediate factor produced by summing over $s_1$: $m_{s_1}(i_1,t_1) = \sum_{s_1} f_{s_1}(s_1) f_{i_1}(i_1,s_1,t_1)$ where $f_{s_1}$ is the first multiplicand and $f_{i_1}$ is the second based on the edge labels. The labels on internal vertices in Figure 2(a) denote how arguments across its parents overlap. We illustrate how the labels for the internal vertices were created by showing the construction for $\mathscr{L}_V(m_{s_1})$:

- *assign each random variable an id*: $s_1 = 1, i_1 = 2, t_1 = 3$

- *begin constructing the label by going through each parent's arguments list and forming a tuple composed of the arguments' ids assigned in the previous step*: since $f_{s_1}$ is the first multiplicand and $f_{i_1}$ the second, we form our label by concatenating "[1]" with "[2,1,3]",

- *add the id of the random variable being summed to the end of the string*: append the string "1" to our label.

---

[*]This is an example of a graphical model that may be constructed during query evaluation over probabilistic databases [18].

Thus, the complete label for $m_{s_1}$ is "{[1], [2,1,3], 1}" (Figure 2(a)). For this example, we are interested in computing the marginals for $i_1, i_2$ and $i_3$ and these marginals are depicted by the leaf vertices.

The main goal of lifted inference is to avoid computing shared factors repeatedly; instead each shared factor should be computed once, and reused whenever required. For instance, in Figure 2(a), we observe that $m_{s_1} \cong m_{s_2}$ since their parents form pairs of shared factors $f_{s_1} \cong f_{s_2}$ and $f_{i_1} \cong f_{i_2}$. This is where the rv-elim graph is useful, it helps us determine the intermediate shared factors generated during inference *before we actually compute them*.

**Property 2.1.** *Vertices $v_1, v_2$ in rv-elim graph $G$ represent shared factors, denoted $v_1 \cong v_2$, (i.e., $f_{v_1} \cong f_{v_2}$ where $f_v$ denotes the factor represented by v), iff:*

- $\forall u_1 \xrightarrow{i} v_1, \exists u_2 \xrightarrow{i} v_2$ *s.t. $f_{u_1} \cong f_{u_2}$ and vice versa (the parents are pairwise shared, in order).*

- $\mathscr{L}_V(v_1) = \mathscr{L}_V(v_2)$ *(arguments overlap info. matches).*

In [18] we observed that by adapting the graph-theoretic notion of *bisimulation* [11], one can determine the equivalence class partitioning of the vertices. Applying the bisimulation algorithm to an rv-elim graph partitions the set of vertices into blocks $B_j$ such that $\forall v_1, v_2 \in B_j, f_{v_1} \cong f_{v_2}$. Once we have such a partition, it is easy to construct a compressed version of the rv-elim graph where each block $B_j$ is represented by a vertex and we introduce an edge $B_j \xrightarrow{i} B_{j'}$ if $\exists v_1 \in B_j, \exists v_2 \in B_{j'}$ s.t. $v_1 \xrightarrow{i} v_2$. Figure 2(b) shows the compressed rv-elim graph constructed from Figure 2(a) where we denote the blocks in square braces next to each vertex; for instance, both $m_{s_1}$ and $m_{s_2}$ have been collapsed to the same vertex $D$. The compressed graph can then be used to perform inference efficiently.

One caveat about the above approach is that since factor

---

**Algorithm 1**: Exact Bisimulation [18]

$d(v) = \begin{cases} 0, \text{if } v \text{ is a root} \\ 1 + \max\{d(v')|v' \to v \in E\} \end{cases}$ /* compute depths */

$\rho \leftarrow \max\{d(v)|v \in V\}$

$B_{0,l} = \{v|v \text{ is root} \wedge \mathscr{L}_V(v) = l\}$ /* compute initial partition */

$C = \{B_{0,l}\}$

$B_i = \{v|d(v) = i\}, \forall i = 1 \ldots \rho$

**for** $i = 1 \ldots \rho$ **do**

    **foreach** $v \in B_i$ **do**    /* construct keys to partition on */

        order parents by block-ids

        construct label $\mathscr{L}_V(v)$

        construct key $k_v$ with $\mathscr{L}_V(v)$ and parents' blocks-ids

    add $B_{i,k} = \{v \in B_i|k_v = k\}$ to $C$

**return** $C$

---

multiplication is a commutative operation, the edge and the internal vertex labels, both of which depend on the order in which the factors are multiplied, can be dynamically altered by choosing a different order. In [18], we proposed ordering the parents of a node using their assigned block-ids before assigning it to a block. This may lead to more symmetry and compression. Algorithm 1 depicts the complete bisimulation algorithm.

Finally to choose the initial elimination order to generate the rv-elim graph, we run a bisimulation on the probabilistic model itself (vertices denote random variables, edges denote dependencies) to compress it. We then run a modified min-size heuristic [13] on the compressed graph, and replace the vertices with the corresponding sets of random variables to get an elimination order for inference.

## 3 Approximate Lifted Inference

While the above approach to performing exact lifted inference can provide significant speedups when the probabilistic model contains moderate to large amounts of symmetry, in many cases we can do much better if we are willing to accept approximations in the marginal probability distributions computed. The main idea here is to explore looser versions of Property 2.1 so that we can partition the vertices of the rv-elim graph into bigger blocks and thus arrive at a smaller compressed rv-elim graph. In what follows, we describe two separate and orthogonal generalizations of Property 2.1 that can be used to implement approximate lifted inference. After that, we discuss how to combine our techniques with bounded complexity inference algorithms and finally, we discuss how to combine all of our proposed ideas together into one approximate lifted inference engine.

### 3.1 Lifted Inference with Approximate Bisimulation

To introduce our first technique we require some notation. Given a vertex, edge labeled graph $G = (V, E, \mathscr{L}_V, \mathscr{L}_E)$ such as an rv-elim graph, let $v_0, \ldots v_n$ denote an $n$-length *vertex path* such that $\forall i = 0, \ldots n : v_i \in V$ and $\forall i = 0, \ldots, n -$ $1 : \exists j$ s.t. $v_i \xrightarrow{j} v_{i+1} \in E$. Further, we say that *label path* or simply, *path*, $l_0(l'_0)l_1(l'_1)\ldots l_n(l'_n)l_{n+1}$ *matches* vertex path $v_0,\ldots v_{n+1}$ (and vice versa) if $\forall i = 0,\ldots,n+1 : \mathscr{L}_V(v_i) = l_i$ and $\forall i = 0,\ldots n : \mathscr{L}_E(v_i \to v_{i+1}) = l'_i$.

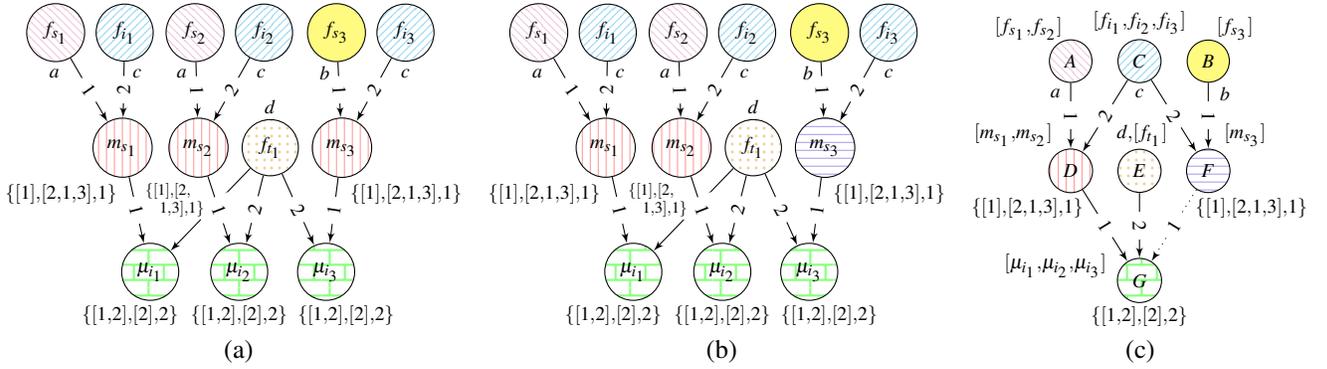

Figure 3: Results of running approximate bisimulation on the running example (shading indicates partitioning), (a) path-length=0 (partitioning on labels, vertices partitioned into 6 blocks) (b) path-length=1 (vertices partitioned into 7 blocks), (c) the compressed graph obtained at path-length=1.

We will now revisit Property 2.1 and try to assign it a path-based interpretation. Using a simple induction (and the fact that edges with the same head have distinct edge labels) it is possible to show that two vertices $v_1$ and $v_2$ in an rv-elim graph are bisimilar *iff their incoming set of paths from the roots are identical*. For instance in Figure 2(a) recall that $m_{s_1} \cong m_{s_2}$, which have the same set of incoming paths from the roots $\{"a(1)\{[1],[2,1,3],1\}", "c(2)\{[1],[2,1,3],1\}"\}$, the matching vertex paths for $m_{s_1}$ are $f_{s_1},m_{s_1}$ and $f_{i_1},m_{s_1}$, resp., and the matching vertex paths for $m_{s_2}$ are $f_{s_2},m_{s_2}$ and $f_{i_2},m_{s_2}$, resp. Notice that this path-based interpretation of Property 2.1 shows that it is a fairly stringent criteria (albeit necessary for exact inference). For instance, consider a case when two vertices deep in the rv-elim graph have large sets of long incoming paths and both sets are almost identical except for one incoming path to the second vertex which has that one label that does not allow it to match any incoming path to the first vertex; based on Property 2.1 these two vertices would be placed in different blocks of the final partition and the compressed rv-elim graph would be correspondingly bloated. This sort of behaviour is, in fact, on display in our running example where $\mu_{i_2} \not\cong \mu_{i_3}$ simply because, of the three incoming paths to $\mu_{i_3}$, $"b(1)\{[1],[2,1,3],1\}(1)\{[1,2],[2],2\}"$ (matching $f_{s_3},m_{s_3},\mu_{i_3}$) doesn't match any of $\mu_{i_2}$'s incoming paths.

Instead of comparing sets of all incoming paths to vertices, we propose to relax Property 2.1 by comparing sets of only $k$-length (and less than $k$-length) incoming paths, where $k$ is a tunable parameter we refer to as the *path-length*. Our compression algorithm permits high compression when the path-length is set to a low value and approaches exact bisimulation when we increase it. Figure 3(a) shows the result of partitioning vertices in our example rv-elim graph with $k$ set to 0 where we simply partition vertices based on their labels. Figure 3(b) is more interesting where we have set $k$ to 1 and so, compare incoming paths of length 1. Now how, in this case, $m_{s_3}$ has been differentiated from $m_{s_1}$ and $m_{s_2}$ since $m_{s_3}$ has an incoming path $"b(1)\{[1],[2,1,3],1\}"$ (matching $f_{s_3},m_{s_3}$) of length 1 which doesn't match any incoming 1-length path of $m_{s_1}$ or $m_{s_2}$. In contrast, $m_{s_1}$, $m_{s_2}$ and $m_{s_3}$ were all placed into the same block in Figure 3(a). Also notice that, in Figure 3(b), $\mu_{i_1}$, $\mu_{i_2}$ and $\mu_{i_3}$ are still partitioned into the same block (leaf vertices tiled with green bricks) and this is because the only path that differentiates $\mu_{i_3}$ from $\mu_{i_1}$ and $\mu_{i_2}$ is a path of length 2 (vertex path $f_{s_3},m_{s_3},\mu_{i_3}$) which is beyond the scope of the current path-length setting of 1. This changes however, when we set path-length to 2 and obtain the results of exact bisimulation shown earlier in Figure 2(a).

The partitioning based on comparing incoming $k$-length paths can be obtained by computing $k$-bisimilarity [11] (for which algorithms are available) since these two properties are equivalent (this can be proved by induction). We formalize the $k$-bisimilarity property as follows:

**Property 3.1.** *Given an rv-elim graph $G = (V,E,\mathscr{L}_V,\mathscr{L}_E)$, $\cong^k$ is defined inductively. For vertices $v_1,v_2 \in V$,*

- $v_1 \cong^0 v_2$ *iff* $\mathscr{L}_V(v_1) = \mathscr{L}_V(v_2)$.

- $v_1 \cong^k v_2$ $(k > 0)$ *iff* $\mathscr{L}_V(v_1) = \mathscr{L}_V(v_2)$ *and* $\forall u_1 \xrightarrow{i} v_1, \exists u_2 \xrightarrow{i} v_2$ *s.t.* $u_1 \cong^{k-1} u_2$ *and vice versa.*

The algorithm for obtaining the partition based on $\cong^k$, Algorithm 2, begins by computing the depth of each vertex $d(v)$ and constructing an initial partition based on labels of the roots and the depths of internal vertices. Throughout Algorithm 2, we maintain two partitions, $X$ and $C$. In the $i^{th}$ iteration, $X$ maintains $\cong^{i-1}$ and is used to update $C$ where we construct $\cong^i$. Note that the inner two loops can be performed in $O(|E|\log D + |V|)$ time (not counting the time spent to construct the vertex labels), where $D$ is the maximum in-degree in the rv-elim graph. Thus, Algorithm 2 runs in $O(k(|E|\log D + |V|))$ time (in contrast to Algorithm 1 which runs in $O(|E|\log D + |V|)$ time). Note that constructing the compressed rv-elim graph corresponding to $\cong^k$ is a bit more complicated now since we are no longer guaranteed that, if two internal vertices fall into the same block of the partition, then the parents will also have been placed into the same block (which holds for Property 2.1). Figure 3(c) (compressed graph obtained at k=1) illustrates this issue where all $\mu$'s have been merged into one block but their $1^{st}$ parents are not, thus $G$ has two $1^{st}$ parents $D$ and $F$ which is problematic if we want to use the compressed graph for inference. Here, we simply get rid of the edge that corresponds to the smaller sized block (the dotted edge $F \rightarrow G$ in Figure 3(c) since $F$ represents a block of size 1 versus $D$ whose block size is 2) to maximize the number of correct marginal probability computations.

### 3.2 Lifted Inference with Factor Binning

We now introduce another way of implementing approximated lifted inference using an orthogonal generalization of Property 2.1. We begin by associating with Property 2.1 a distance-based interpretation. Recall that, Property 2.1 bins two factors into the same block of the partition when we can guarantee that their input-output mappings are exactly the same without actually computing them. Stated differently, given any user-defined distance measure that can measure the "distance" between two factors, Property 2.1 deems that these factors belong to the same block only if the distance between them is zero. Note that the converse is not true. That is, it is possible for two internal vertices in the rv-elim graph to actually represent factors that comprise of identical input-output mappings but because their parents do not belong to the same blocks or because the parents' arguments don't overlap in the same fashion, Property 2.1 cannot bin these into the same block of the partition. We illustrate this with the following example:

$$\sum_Y \left( \begin{array}{cc|c} X & Y & f_1 \\ \hline \texttt{t} & \texttt{t} & 0.8 \\ \texttt{t} & \texttt{f} & 0.2 \\ \texttt{f} & \texttt{t} & 0.4 \\ \texttt{f} & \texttt{f} & 0.6 \end{array} \times \begin{array}{c|c} Y & f_2 \\ \hline \texttt{t} & 0.5 \\ \texttt{f} & 0.5 \end{array} \right) = \begin{array}{c|c} X & m_Y \\ \hline \texttt{t} & 0.5 \\ \texttt{f} & 0.5 \end{array}$$

$$\sum_{Y'} \left( \begin{array}{cc|c} X' & Y' & f_1' \\ \hline \texttt{t} & \texttt{t} & 0.2 \\ \texttt{t} & \texttt{f} & 0.8 \\ \texttt{f} & \texttt{t} & 0.6 \\ \texttt{f} & \texttt{f} & 0.4 \end{array} \times \begin{array}{c|c} Y' & f_2' \\ \hline \texttt{t} & 0.5 \\ \texttt{f} & 0.5 \end{array} \right) = \begin{array}{c|c} X' & m_{Y'} \\ \hline \texttt{t} & 0.5 \\ \texttt{f} & 0.5 \end{array}$$

where $\texttt{t}$ and $\texttt{f}$ denote $\texttt{true}$ and $\texttt{false}$ resp. Notice how factors $f_1$ and $f_1'$ have different input-output mappings ($f_1(\texttt{t},\texttt{t}) = 0.8 \neq 0.2 = f_1'(\texttt{t},\texttt{t})$) and hence cannot be binned into the same block which means that it is not possible to determine that the resulting factors $m_Y$ and $m_{Y'}$ comprise of the same input-output mappings solely using Property 2.1. This, in turn, means that any intermediate factors derived from these two factors during the inference process will always be binned separately, thus leading to a

---

**Algorithm 2**: Approximate Bisimulation(k)

$d(v) = \begin{cases} 0, \text{if } v \text{ is a root} \\ 1 + \max\{d(v')|v' \rightarrow v \in E\} \end{cases}$
$\rho \leftarrow \max\{d(v)|v \in V\}$
$B_{0,l} = \{v|d(v) = 0 \wedge \mathscr{L}_V(v) = l\}$
$B_i = \{v|d(v) = i\} \forall i = 1 \dots \rho$
$C \leftarrow \{B_{0,l}\}_{\forall l} \cup \{B_i\}_{i=1}^{\rho}$
$X \leftarrow C$
**for** $j = 1 \dots k$ **do**
    **for** $i = 1 \dots \rho$ **do**
        **foreach** $B \in C$ at depth $i$ **do**
            order parents by block-ids in $X$
            construct labels $\mathscr{L}_V(v) \forall v \in B$
            construct key $k_v \forall v \in B$ with $\mathscr{L}_V(v)$, parent block-ids in $X$
            partition $B$ based on keys $k_v$
            replace $B$ in $C$ with new blocks
    $X \leftarrow C$
**return** $C$

---

bloated compressed rv-elim graph.

Such symmetries can not be captured without actually looking into the factors and computing the distance between them (any distance measure such as KL-divergence or root mean squared distance would do). For this purpose, we ask the user for a separate parameter $\varepsilon$, that specifies an upper bound on the distance between two factors for them to be considered shared. Note that, unlike the previous algorithm, we can not compute distance between two intermediate factors without computing the factors.

To determine such a distance-based partitioning of the factors, we will need to solve the *factor binning* problem (FB):

Given:      set of factors $\mathscr{F} = \{f_1, \dots f_n\}$
              threshold $\varepsilon$, distance function $\text{dist}(\cdot, \cdot)$
Return:     $\text{argmin}_{\mathbf{F} \subseteq \mathscr{F}} |\mathbf{F}|$
such that   $\forall f_i \in \mathscr{F} \setminus \mathbf{F} \; \exists f \in \mathbf{F} \text{ s.t. } \text{dist}(f_i, f) \leq \varepsilon$

We will shortly show that the factor binning problem is equivalent to the *dominating set* problem (DS):

Given:      graph $\mathbf{G}$ with vertex set $\mathbf{V}$ and edge set $\mathbf{E}$
              denote by $N_v$ neighborhood of vertex $v$
Return:     $\text{argmin}_{\mathbf{D} \subseteq \mathbf{V}} |\mathbf{D}|$
such that   $\forall v_i \in \mathbf{V} \setminus \mathbf{D} \; \exists v \in \mathbf{D} \text{ s.t. } v \in N_{v_i}$

**Theorem 3.2.** *FB is equivalent to DS.*

*Proof.* The proof is in two parts, we first show that any instance of FB can be reduced to DS and vice versa. To show the first part, we specify the reduction to DS. Given an instance of FB, define the corresponding DS by setting:

$$DS_{FB} : \mathbf{V} = \mathscr{F}, N_{f_i} = \{f_i\} \cup \{f | \text{dist}(f_i, f) \leq \varepsilon\}$$

---

**Algorithm 3:** Factor Binning($\varepsilon$)

$d(v) = \begin{cases} 0, \text{if } v \text{ is a root} \\ 1 + \max\{d(v')|v' \to v \in E\} \end{cases}$

$\rho \leftarrow \max\{d(v)|v \in V\}$

$B_{0,l} = \{v|v \text{ is a root} \wedge \mathcal{L}_V(v) = l\}$

**FB**   instantiate one factor per block $B_{0,l}$

**FB**   $\mathbf{B}_0^{hs} \leftarrow$ compute hitting set and construct new set of blocks by merging $\{B_{0,l}\}$

$C = \mathbf{B}_0^{hs}$

$B_i = \{v|d(v) = i\}, \forall i = 1 \ldots \rho$

**for** $i = 1 \ldots \rho$ **do**

  **foreach** $v \in B_i$ **do**

    order parents by block-ids

    construct label $\mathcal{L}_V(v)$

    construct key $k_v$ with $\mathcal{L}_V(v)$ and parents' blocks-ids

  $B_{i,k} = \{v \in B_i|k_v = k\}$

**FB**   instantiate one factor per new block $B_{i,k}$

**FB**   $\mathbf{B}_i^{hs} \leftarrow$ compute hitting set and construct new set of blocks by merging $\{B_{i,k}\}$

  $C \leftarrow C \cup \mathbf{B}_i^{hs}$

**return** $C$

---

Note that any solution to $DS_{FB}$ is a solution to FB. We show this by contradiction. Suppose solution $\mathbf{D}$ to $DS_{FB}$ is not a solution to FB, i.o.w., $\exists f_i \in \mathscr{F} \setminus \mathbf{D}$ s.t. $\text{dist}(f_i, f) > \varepsilon$, $\forall f \in \mathbf{D}$. This implies $N_{f_i} \cap \mathbf{D} = \emptyset$ which means that $\mathbf{D}$ is not a solution to $DS_{FB}$ and thus we have a contradiction. Similarly, any solution to FB is a solution to $DS_{FB}$. Again, assume that solution $\mathbf{F}$ to FB is not a solution to $DS_{FB}$. Thus, $\exists f_i \in \mathscr{F} \setminus \mathbf{F}$ s.t. $N_{f_i} \cap \mathbf{F} = \emptyset$. This implies $\text{dist}(f_i, f) > \varepsilon$, $\forall f \in \mathbf{F}$ which means $\mathbf{F}$ is not a solution to FB and we have a contradiction. Given that solution spaces of FB and $DS_{FB}$ are same, and that the objective functions are also same, we have shown that FB can be solved by solving $DS_{FB}$.

The reduction in the other direction is also easy. Given an instance of DS, define the corresponding $FB_{DS}$ by setting:

$$FB_{DS}: \quad \mathscr{F} = \mathbf{V}, \varepsilon = 0$$
$$\text{dist}(v_i, v_j) = \begin{cases} 0 & \text{if } (v_i, v_j) \in \mathbf{E} \\ 1 & \text{otherwise} \end{cases}$$

It's easy to show that DS, $FB_{DS}$ share the same soln. space.
$\qquad\qquad\qquad\qquad\qquad\qquad\qquad\qquad\qquad\qquad\square$

DS is NP-Complete [7]. Further, Feige [6] showed that DS is not approximable to a factor of $(1 - o(1))ln(|\mathbf{V}|)$ unless NP has "slightly super-polynomial time" algorithms (or $NP \subset DTIME(n^{\log(\log(|\mathbf{V}|))})$). One way to solve DS is to utilize the fact that it is a special case of *set cover* and use the obvious greedy heuristic (described below) for set cover. This gives us an $ln(|\mathbf{V}|)$-approximation algorithm [21]. Thus, for our experiments we use the same greedy approach to solve FB. FB is also equivalent to the $\rho$-*dominating set* problem [2], which, in turn, is the converse of the classic *k-center* problem [12] where we are given a graph from which we need to choose a subset of $k$ vertices so that their distance from the other vertices is minimized. Note that, when the distance function satisfies special properties, better algorithms may be available. For instance, for euclidean spaces, near-optimal factor binning is possible [10], especially when the factor sizes are not large.

The algorithm to obtain the greedy solution for FB is to first construct each subset $N_{f_i}$ (as defined above) and repeatedly pick $f_i$ corresponding to the current largest $N_{f_i}$ to include into our solution. Every time we pick $f_i$, we update all $N_{f_j}$'s by deleting from them all factors that are within $\varepsilon$ distance of $f_i$. Another question we need to consider is whether to bin factors based on distance once and then run approximate lifted inference or whether to bin the intermediate factors based on distance also. For our experiments, we also binned the intermediate factors since this allows us to compress the rv-elim graph more aggressively. Algorithm 3 shows the complete algorithm to run approximate lifted inference using FB. Algorithm 3 is essentially Algorithm 1 with extra lines for FB computations (marked **FB**).

## 3.3 Bounded Complexity Lifted Inference

The approximation techniques we have introduced so far do not alleviate the worst-case complexity of the inference procedure. In other words, these techniques would not help if the ground inference procedure is associated with high treewidth (common with structured probabilistic graphical models). Next we show how to incorporate the mini-bucket scheme [5], a bounded complexity approximate (ground) inference algorithm, with our ideas. This allows us to keep a tight control over the complexity of inference incurred.

The mini-buckets scheme is a modification of the variable elimination algorithm [23] where at each step instead of eliminating a random variable by multiplying *all* factors it appears as argument in, one devises a set of *mini-buckets* each containing a (disjoint) subset of factors that contains that variable as argument and then eliminates the variable separately from each mini-bucket. More precisely, given a set of factors, one first constructs a canonical partition such that all subsumed factors are placed into the same bucket of the partition. A factor $f$ is said to be subsumed by factor $f'$ if any argument of $f$ is also an argument of $f'$. After constructing the canonical partition, the user has two choices:

- construct mini-buckets by restricting the total number of arguments $i$ (a user-defined parameter) in each mini-bucket. Since inference complexity is directly affected by the size of the largest factor encountered, this is one way to control the amount of computation incurred.

- specify how many buckets $m$ of the canonical partition to merge to form a mini-bucket. Again, this (indirectly) controls the size of the largest factor generated and keeps the complexity bounded.

Dechter and Rish [5] show how such a modification of the variable elimination algorithm provides an upper bound over the numbers produced in the resulting factors.

It is easy to combine our approaches with the mini-bucket scheme. Instead of building the rv-elim graph by introducing internal vertices corresponding to intermediate factors produced by multiplying all factors involving a certain random variable as argument, we simply introduce vertices corresponding to factors produced by the mini-bucket scheme. Since our approaches work on any rv-elim graph, this requires no change to the approaches presented earlier, while keeping the complexity of inference bounded.

### 3.4 Unified Lifted Inference Engine

By interleaving the various steps, it is possible to combine all the ideas we have presented in this section into one unified approximate lifted inference engine. Our combined inference engine takes a set of eight parameters which define the combinations of techniques we would like to invoke (see Table 1). The experiments presented in the next section use this generic inference engine.

## 4 Experimental Evaluation

We conducted experiments on synthetic and real data to determine how lifted inference with approximate bisimulation and factor binning perform on their own. We also report experiments with our unified lifted inference engine where we used both approaches in tandem. Each number we report is an average over 3 runs, our implementation is in JAVA and our experiments were performed on a machine with a 3GHz Xeon processor and 3GB RAM. We compare our results with two baseline algorithms: A ground inference procedure which is basically variable elimination [23] modified so that we obtain all marginals in a single pass, and the exact lifted inference procedure reviewed in Section 2. We report two metrics for each experiment: run times incurred by the various algorithms in seconds (*Time*) and error measured by computing the average number of marginal probabilities which were not within $10^{-8}$ of their correct values (*Avg. #Probs. Incorrect*).

### 4.1 Synthetic Bayesian Network Generator

We set up a synthetic Bayesian network (BN) generator to test various aspects of our algorithms. The generator produces BNs where the random variables are organized in layers and random variables from the $i^{th}$ layer randomly choose parents from the $i - 1^{th}$ layer. For our experiments, we generated BNs with 3 layers: $1^{st}$ layer contained 1000 random variables, $2^{nd}$ 500 and $3^{rd}$ 250. We introduced priors randomly for each variable in the first layer, every $25^{th}$ prior was identical. The random variables in the last layer are our query variables for which we computed marginal probabilities. All random variables had domain of size 30. To generate factors defining the dependency between random variables from the $i^{th}$ and $i - 1^{th}$ layers, for each variable in the $i^{th}$ layer, we randomly chose 2 parents from the previous layer. Two children can choose the same parents, so we generated non-tree structured BNs. All factors with children from the $i^{th}$ layer are identical. This closely follows most structured probabilistic graphical models we have come across, where the priors usually closely resemble each other but may not be identical; whereas the factors defining dependencies between various random variables come from generic rules and are thus identical. We used a parameter to control how many times a random variable can be picked as a parent. This helps vary the complexity of the inference problem. We also used a parameter to add random noise after the factors are generated. We tried other parameter settings as well and the trends were as expected. For instance, increasing domain size increases the speedups obtained since with larger domains, we increase the time spent summing over random variables and multiplying factors while running ground inference – our lifted inference procedures are designed to save on this assuming the symmetry among factors is kept constant. Similarly, increasing the number of random variables with constant symmetry also increases speedups obtained.

| Parameter Name: Description |
|---|
| UB (bisimulation): compresses rv-elim graphs if `true` |
| PL (path length): approximate bisimulation parameter, use exact bisimulation when set to $\infty$ |
| $\varepsilon$: factor binning parameter, uses factor binning if $\varepsilon > 0$ |
| UMB (mini-bucket): allows using mini-buckets if `true` |
| ACR (arg. count restriction): if `true` then restricts based on number of arguments in mini-buckets |
| MBR (mini bucket restriction): if ACR=`true` then this is $i$ (the max number of args per mini-bucket), else it is interpreted as $m$ (the number of canonical partition buckets merged to form a mini-bucket). |

Table 1: Parameters for our unified lifted inference engine.

### 4.2 Lifted Inference with Approximate Bisimulation

Our first set of experiments tests our algorithm for lifted inference with approximate bisimulation. The results are reported in Figure 4 (a) and Figure 4 (d). The plots show that as we increase path-length (x-axis in these plots) the time for inference (Figure 4 (a)) slowly increases but error decreases (Figure 4 (d)). The solid line with triangles depict the results of running lifted inference with approximate bisimulation without mini-buckets, and with path-length set to 3 we see that the error stands around 18%; the inference procedure took about 3 seconds to run, which is almost a 3 times speedup over exact lifted inference (which took 8.2 seconds) and almost a 9 times speedup over ground inference (which took 25.95 sec). All the other lines in the plots correspond to lifted inference with approximate bisimulation run with various mini-bucket schemes. Among these, the mini-bucket scheme with mini-buckets restricted by argument count at $i = 3$ seems to be a promising setting (dotted line with triangles) since it runs faster than lifted inference with approximate bisimulation but does not incur significantly higher error. Another interesting thing that shows up in these plots is that with mini-buckets with $i = 4$ or $m = 2$ at path-length set to 3, the time taken to run inference goes up noticeably. This shows that at very low path-lengths, using mini-buckets could actually lead to loss of symmetry in the rv-elim graph.

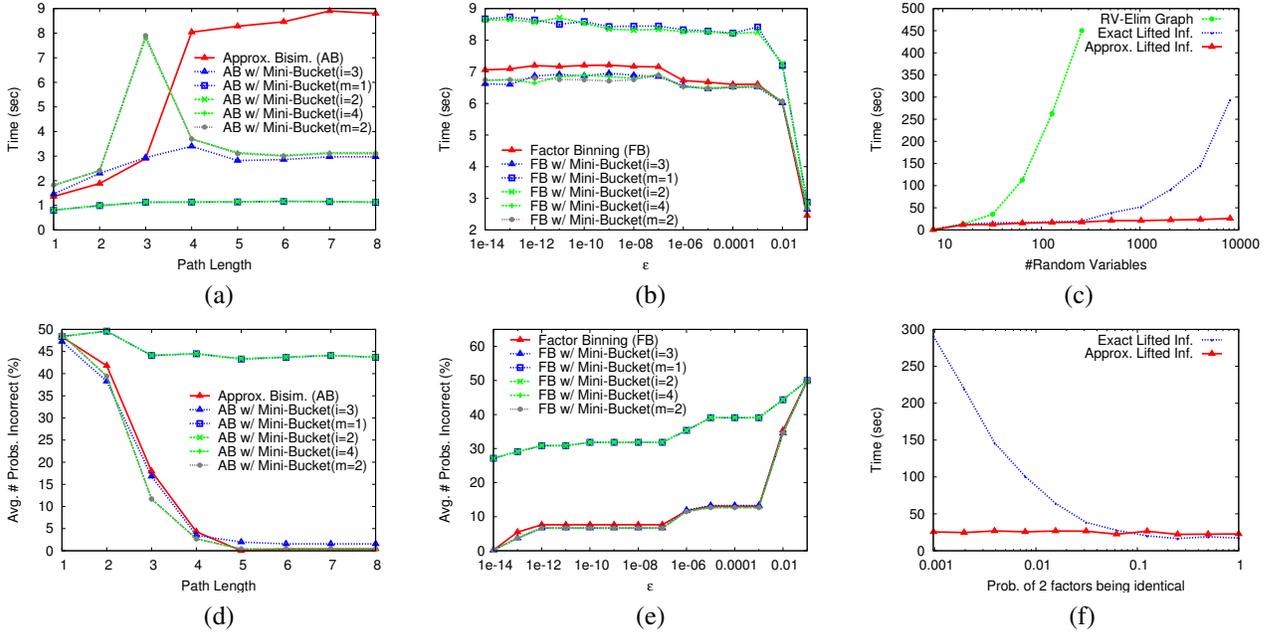

Figure 4: Experimental results for the various inference algorithms. (a) and (d) report time and error for lifted inference with approximate bisimulation (var. elim. took 25.95 sec and exact lifted inference took 8.2 sec). (b) and (e) report time and error for lifted inference with factor binning (var. elim. took 33.12 sec and exact lifted inference took 25.24 sec). (c) and (f) report time and error for the unified lifted inference engine.

### 4.3 Lifted Inference with Factor Binning

Our second set of experiments tests our factor binning approach. The results are shown in Figure 4 (b) and Figure 4 (e). For these experiments, we used root mean squared distance to compare two factors. More precisely, given two factors $f_1$ and $f_2$ with a common joint domain $D$, $\text{dist}(f_1, f_2) = \sqrt{\frac{1}{|D|} \sum_{\mathbf{x} \in D} (f_1(\mathbf{x}) - f_2(\mathbf{x}))^2}$. The plots show that as we increase $\varepsilon$ (on the x-axis) the times for inference go down (Figure 4 (b)), and the error goes up (Figure 4 (e)). On these experiments, ground inference took about 33 seconds and exact lifted inference took 25.24 seconds which means factor binning without mini-buckets (solid line with triangles) achieves a speedup of about 3.5 times over exact lifted inference and a speedup of almost 5 times over ground inference. Among the various mini-bucket schemes, once again $i = 3$ (dotted line with triangles) seems to be the best setting which gives small but noticeable reductions in run-times at almost no cost to accuracy. Notice that mini-buckets with small settings of either $m$ or $i$ tends to perform very poorly neither giving good accuracies nor providing good run-times and this is likely due to the sheer number of factors with which we are dealing. At such small settings, the mini-bucket scheme produces a lot of factors and computing the hitting set (which has a quadratic time complexity) becomes too expensive.

### 4.4 Unified Lifted Inference Engine

In our last set of experiments, we used both approximate bisimulation (path-length=3) and factor binning ($\varepsilon = 0.01$) with mini-buckets (restricted by argument count $i = 3$). Here we report run-times for probabilistic models with varying number of random variables. The results are reported in Figure 4 (c) and Figure 4 (f). As should be clear from Figure 4 (c), with increasing size of the probabilistic model all three inference procedures, ground inference, exact lifted inference and approximate lifted inference, show an increase in run-time but there is an order of magnitude difference in times between ground inference and exact lifted inference (which partitions identical factors together) and another order of magnitude speedup over exact lifted inference for approximate lifted inference (which also bins nearly identical factors together) while keeping the accuracy within bounds. Thus approximate lifted inference is more than two orders of magnitude faster than ground in-

| Dataset | Inf. Alg. | Time (s) | Arith. Ops. | Rem. Ops. | Acc. |
|---------|-----------|----------|-------------|-----------|------|
| Cora | Ground Inf. | 163.5 | 163 | 0.5 | 77.8 |
|      | Lifted Inf. | 60.6 | 59.9 | 0.7 | 73 |
| CiteSeer | Ground Inf. | 101.0 | 100.8 | 0.2 | 68.7 |
|          | Lifted Inf. | 65.0 | 63.9 | 1.1 | 66.8 |

(a)

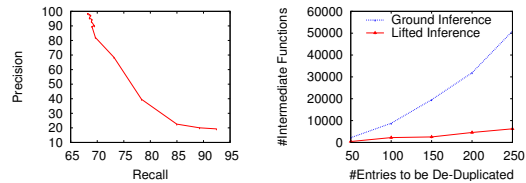

(b)                    (c)

Figure 5: (a) Times for Cora and CiteSeer. (b) Precision-Recall curve for Cora-ER and (c) number of factors generated.

ference. The accuracies for approximate lifted inference for these experiments varied between 65-95%. For these complex networks, we could not run ground inference on models with more than 256 random variables due to memory limitations. Figure 4 (f) makes it clear how the runtime between exact lifted inference and approximate lifted inference varies. Here we set all priors in our probabilistic model similar to each other but varied the probability of two factors being identical to each other. The plot shows that as this probability increases, exact lifted inference captures the symmetry and does better, whereas approximate lifted inference keeps run-times low throughout.

### 4.5 Experiments on Real-World Data

We experimented with a number of real world datasets. We first report results on the Cora [14] and CiteSeer [9] datasets. The Cora dataset contains 2708 machine learning papers with 5429 citations; each paper is labeled from one of seven topics. The CiteSeer dataset consists of 3316 publications with 4591 citations; each paper is labeled with one of 6 topics. The task is to predict the correct topic label of the papers. We divided each dataset into three roughly equal splits and performed three-fold cross valiation. For each experiment, we train on two splits and test on the third, randomly choosing 10% of the papers' class labels to be our query nodes. Each number we report is an average across all splits. Note that, for these experiments, using the citations in the datasets we produce Markov networks with unbounded treewidth and then perform *collective classification* [19], so we compare against ground inference with mini-buckets restricted to 6 arguments. Also, while testing on the third split, we include as evidence topic labels of the papers belonging to the training set linked to from the test set. We tried various parameter settings with our approximate lifted inference engine and report the best results. As Table 5 (a) shows, we obtained a 2.7 times speedup for Cora and 1.55 times speedup for CiteSeer with our approximate lifted inference engine over ground inference. The loss in accuracy was 4.8% for Cora and 1.9% for CiteSeer. These results were obtained with path length = 2, $\varepsilon = 0.01$ and using mini-buckets restricted to 6 arguments. We also show how much time was spent by each inference scheme to multiply factors and sum over random variables (arithmetic operations or "Arith. Ops." in Table 5 (a)) and the remaining operations (or "Rem. Ops." in Table 5 (a)). As should be clear from Table 5 (a), the various bisimulation algorithms and hitting set computations do not really add much overhead on these datasets; we spend about 0.2 seconds, for Cora, and 0.9 seconds, for CiteSeer, more than ground inference to implement lifted inference.

We also experimented with the Cora dataset for entity resolution (Cora-ER) [1]. For this experiment, we used a Markov logic network with 46 distinct rules. Unfortunately, we could not get any noticeable speedup for this dataset. This dataset consists solely of random variables with domain size 2 (match/non-match). As a result, all the factors produced are extremely small in size (size of a factor is determined by the number of rows in it) which implies that the time spent performing arithmetic operations (multiplying factors and eliminating random variables) is not the bottleneck during inference. The techniques proposed in this paper are mainly directed towards reducing the time spent to perform arithmetic operations. However, we do present the precision-recall curve we obtained for Cora-ER (Figure 5 (b), increasing argument count restriction for the mini-buckets scheme reduces precision but increases recall) and we also counted the number of intermediate factors computed by ground and lifted inference for various samplings of the dataset consisting of 50-250 bibliographic citations to be deduplicated. Figure 5 (c) shows that lifted inference produces far fewer intermediate factors during inference than ground inference; recall that ground inference produces an intermediate factor everytime a random variable is eliminated but lifted inference saves on this computation by computing one factor for each *block* in the final partitioning. This, in turn, indicates that the dataset possesses symmetry which could lead to speedups if the domain sizes of the random variables and factors were large. Note that Figure 5 (c) also gives an idea of the reduced memory consumption for lifted inference.

## 5 Related Work

Poole [17] was one of the first to show that variable elimination [23] can be modified to directly work with first-order representations of random variables and factors (or clique potentials) to avoid propositionalization. Subsequently, de Salvo Braz et al. [3] further developed on Poole's work and referred to it as *inversion elimination*. They also introduce another technique for lifted inference known as *counting elimination* which is more expensive than inversion elimination but can help in certain situations where the ground model's treewidth renders ground inference infeasible. It is straightforward to show that the bisimulation approach to lifted inference subsumes inversion elimination (and partial inversion [4]). Given a computation of the form $\sum_Y \sum_{X_i} \prod_i \psi(X_i, Y)$ (all $\psi$'s are shared factors), inversion elimination avoids the complexity of eliminating each $X_i, \forall i = 1, \ldots n$ separately by pushing each summation of $X_i$ against the corresponding $\psi$, eliminating $X_i$ once and then eliminating $Y$: $\sum_Y \sum_{X_i} \prod_{i=1}^n \psi(X_i, Y) = \sum_Y \prod_{i=1}^n \sum_{X_i} \psi(X_i, Y) = \sum_Y \prod_{i=1}^n \psi'(Y) = \sum_Y \psi'^n(Y) = \psi''()$. Figure 6 shows how our approach achieves the same.

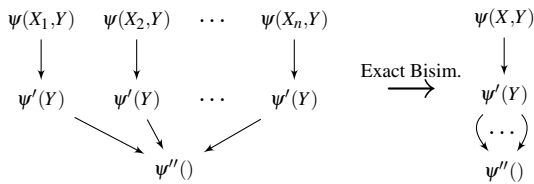

Figure 6: Inversion elimination is a special case of bisimulation-based inference: the rv-elim graph and its compressed version. Labels not shown for legibility.

In other related work, Singla and Domingos [20] propose an approach where they run a bisimulation-like algorithm on the factor graph representing the probabilistic model to find clusters of random variables that send and receive identical messages which helps speed up inference with loopy belief propagation (LBP) [22], a ground approximate inference algorithm. Our approaches differ from theirs on two counts: first, their approach requires as input the specification of the probabilistic model in first-order format (ours, in effect, determines the first-order representation) and second, as the authors acknowledge in their paper, LBP often has problems with convergence, whereas the approach we describe in Section 2 always returns exact marginals and the approach we describe in Section 3, even though it is approximate, is always guaranteed to converge.

## 6 Conclusion and Future Work

In this paper, we described light-weight, generally applicable approximation algorithms for lifted inference based on the graph theoretic concept of bisimulation. Essentially, our techniques are wrap-arounds for variable elimination [23] and can be used whenever variable elimination is applicable, including computing joint conditional probabilities and MAP assignments (by switching from the sum-product operator to max-product). One interesting avenue of future work is to look for other bounded complexity inference algorithms (besides mini-buckets) that can be combined with the techniques introduced in this paper. Other avenues of future work are determining the optimal values of the various parameters (path-length and $\varepsilon$) automatically and building the compressed rv-elim graph directly from the first-order description of the probabilistic model.

## Acknowledgements

This work was supported in part by NSF Grants No. IIS-0438866 and IIS-0546136. We would also like to thank the anonymous reviewers for their comments and suggestions and Parag Singla for sharing with us the Cora-ER MLN.